\documentclass[11pt,letterpaper]{article}
\usepackage{cogsys}
\usepackage[T1]{fontenc}
\usepackage{times}
\usepackage[pdftex]{graphicx} 
\usepackage[cachedir=.]{minted}
\usepackage{multirow}
\usepackage{booktabs}
\usepackage{xcolor}
\usepackage[hidelinks]{hyperref}
\usepackage{csquotes}
\usepackage{amssymb}

\usepackage{caption}
\usepackage{subcaption}
\usepackage{amsmath}

\definecolor{mapA}{HTML}{1F6FB4} 
\definecolor{mapB}{HTML}{C42B2B} 
\definecolor{mapC}{HTML}{1E8A3C} 
\definecolor{mapD}{HTML}{8548A8} 
\definecolor{mapF}{HTML}{C77400} 
\definecolor{mapG}{HTML}{0E8C8C} 
\definecolor{mapH}{HTML}{C0247A} 
\definecolor{mapI}{HTML}{7A4A28} 

\newcommand{\mA}[1]{\textcolor{mapA}{#1}}
\newcommand{\mB}[1]{\textcolor{mapB}{#1}}
\newcommand{\mC}[1]{\textcolor{mapC}{#1}}
\newcommand{\mD}[1]{\textcolor{mapD}{#1}}
\newcommand{\mF}[1]{\textcolor{mapF}{#1}}
\newcommand{\mG}[1]{\textcolor{mapG}{#1}}
\newcommand{\mH}[1]{\textcolor{mapH}{#1}}
\newcommand{\mI}[1]{\textcolor{mapI}{#1}}

\usepackage{natbib}
\cogsysheading{13}{2026}{}{8/2026}{10/2026}

\ShortHeadings{SMTB: Fast Structure-Mapping with Tight Bounds}
              {D.\ Weitekamp}

\begin{document} 

\title{SMTB: Fast Structure-Mapping with Tight Bounds}
 
\author{Daniel Weitekamp}{weitekamp@gatech.edu}
\author{Christopher MacLellan}{cmaclell@gatech.edu}
\address{School of Interactive Computing, Georgia Tech, 
         Atlanta, GA 30332 USA}

\vskip 0.2in
 
\begin{abstract}

Structure-mapping forms analogies by aligning systems of relationally connected elements based on shared structure instead of surface features. We introduce a new structure-mapping algorithm: Structure-Mapping with Tight Bounds (SMTB) that is 5--15x faster than the structure-mapping engine (SME) and about 50\% better at finding mappings in large nested domains. SMTB is part of the broader Cognitive Rule Engine (CRE) project, a flexible multi-language-compatible framework with an accessible Python interface to state-of-the-art C++ implementations of core algorithms commonly used in cognitive systems such as pattern matching, planning, and structure-mapping. CRE and SMTB are designed to work with a wide range of representation choices. Unlike SME, which biases higher-order correspondences in tree-like predicate logic, SMTB maximizes relational connectivity without privileging higher-order relations. This allows SMTB to work just as well over arbitrary relational graphs as it does in tree-like domains of nested predicate logic. We discuss situations where privileging "higher-orderness" in structure-mapping can cause issues, and illustrate how SMTB avoids failure modes that SME would encounter in these situations. We also provide an evaluation comparing SMTB to SME v4 over 5845 domain pairs from the SME corpus.


\end{abstract}

\section{Introduction}

Central to analogical reasoning is the ability to recognize relational similarities between things (e.g., factual information, objects in natural scenes, or mental abstractions) based on their shared relational structure. This is an essential mechanism in human cognition that mediates sense-making and learning \citep{gentner2010bootstrapping} and allows us to reason about new, uncertain situations by mapping them onto known ones. \textit{structure-mapping} \citep{gentner1983structure} is a computational theory of the formation of structural analogies, which characterizes analogies as self-consistent one-to-one mappings between \textit{base} and \textit{target} domains of relationally encoded entities. In cognitive systems, structure-mapping has many uses \citep{gentner2011computational}, including drawing analogies between experiences, aligning knowledge structures to combine or relax them into higher-order generalizations, and doing partial matching to assess the degree of applicability of prior knowledge in new situations. In this work, we present structure-mapping with Tight Bounds (SMTB) and demonstrate areas where it improves upon the structure-mapping engine (SME) \citep{falkenhainer1989structure}---a system that has served as the structure-mapping implementation for 40 years of cognitive systems, including several ongoing projects \citep{gentner2025multidecade}. SME has been used in cognitive systems such as SAGE \citep{kuehne2000seql, mclure2015nearmisses} and CogSketch \citep{forbus2011cogsketch}, in the Companion cognitive architecture \citep{forbus2017companions}, and in studies of analogical reasoning in solving AP physics problems \citep{klenk2009analogical}.

We began the development of SMTB to serve as a fast, Python-accessible structure-mapping implementation for our students working on our Python-based cognitive systems. During development, we found that SME's approach imposed constraints and search biases that explicitly precluded structure-mapping solutions that were desirable in our use cases. These issues, which we expand on in Section 3, motivated SMTB's approach which relaxes some of SME's constraints.    


\begin{table}[t]
\centering
\small
\begin{tabular}{ll}
\toprule
\textbf{Base: water flow} & \textbf{Target: heat flow} \\
\midrule
\texttt{\mG{Greater$_1$}(\mF{Pressure}(\mA{beaker}),} & \texttt{\mG{Greater}(\mF{Temp}(\mA{coffee}),} \\
\texttt{~~~~~~~~~\mI{Pressure}(\mB{vial}))}        & \texttt{~~~~~~~~\mI{Temp}(\mB{icecube}))} \\
\texttt{Greater$_2$(Diameter(\mA{beaker}),}           & \\
\texttt{~~~~~~~~~Diameter(\mB{vial}))}             & \\
\texttt{\mH{Flow}(\mA{beaker}, \mB{vial}, \mC{water}, \mD{pipe})}
                                                  & \texttt{\mH{Flow}(\mA{coffee}, \mB{icecube}, \mC{heat}, \mD{bar})} \\
\texttt{Cause(\mG{Greater$_1$}(...), \mH{Flow}(...))} & \\
\texttt{Liquid(\mC{water})}                       & \texttt{Liquid(\mA{coffee})} \\
\texttt{Flat-Top(\mC{water})}                     & \\
\bottomrule
\end{tabular}
\captionsetup{font={stretch=.9}} 
\caption{
Water-flow / heat-flow analogy \citep{falkenhainer1989structure}. Colors mark the correspondences of the intended mapping between base entities: 
\mA{\texttt{beaker}}$\rightarrow$\mA{\texttt{coffee}}, 
\mB{\texttt{vial}}$\rightarrow$\mB{\texttt{icecube}}, 
\mC{\texttt{water}}$\rightarrow$\mC{\texttt{heat}}, 
\mD{\texttt{pipe}}$\rightarrow$\mD{\texttt{bar}}, 
and expressions: 
\mG{\texttt{Greater$_1$}}$\rightarrow$\mG{\texttt{Greater}}, 
\mH{\texttt{Flow}}$\rightarrow$\mH{\texttt{Flow}},  
\mF{\texttt{Pressure}}$\rightarrow$\mF{\texttt{Temp}}. 
\mG{\texttt{Greater$_1$}} and \mG{\texttt{Greater$_2$}} are distinct instances of \texttt{Greater}, over \texttt{Pressure}, and \texttt{Diameter}. 
}
\vspace{-0.5em}
\label{tab:sm-example}
\end{table}

\subsection{Water Flow $\rightarrow$ Heat Flow Example}

\citet{falkenhainer1989structure} provide the conical structure-mapping example, which draws an analogy between water flow driven by differences in pressure and heat flow driven by differences in temperature (shown in Table \ref{tab:sm-example}). Base entity identifiers (e.g., \texttt{beaker}, \texttt{vial}, \texttt{water}, \texttt{pipe}) are connected to one another through hierarchies of relational expressions encoded as predicates (e.g., \texttt{Greater}, \texttt{Pressure}, \texttt{Diameter}, \texttt{Flow}). The base encodes a situation where "water flows from a beaker to a vial through a pipe, because the pressure in the beaker is greater". In the target: "heat flows from coffee to an ice cube through a bar". A successful mapping between them enables inference of a missing idea in the target: the flow occurs "because the temperature in the coffee is greater than the ice cube". Table \ref{tab:sm-example} is aligned and colored to illustrate the ideal analogical mapping for this case. The ideal mapping skips shallow attribute correspondences, such as \texttt{water} and \texttt{coffee} both being \texttt{Liquid}, and instead identifies stronger structural alignment in higher-order relations. A naive commitment to \texttt{water}$\rightarrow$\texttt{coffee} (because they are both liquid) would have prevented the higher-order \mG{\texttt{Greater$_1$}}$\rightarrow$\mG{\texttt{Greater}}, and \mH{\texttt{Flow}}$\rightarrow$\mH{\texttt{Flow}} mappings. In general, structure-mapping systems such as SME can optionally yield multiple mappings that can differ in their assigned alignment score and ground-truth quality. For instance, \mG{\texttt{Greater$_2$}}$\rightarrow$\mG{\texttt{Greater}} along with \texttt{Diameter}$\rightarrow$\texttt{Temp} is another logically consistent mapping---albiet one that blocks a candidate inference of a cause in the target, and thus may have a lower score.

\subsection{Our Motivations}

\textbf{Computational Models of Learning:}
SMTB was developed as a fast, flexible structure-mapping system for computational models of learning \citep{weitekamp2023computational} such as the Apprentice Learner \citep{maclellan2016apprentice} and AI2T \citep{weitekamp2024ai2t}, which simulate human learning from intelligent tutoring system instruction. Unlike statistical simulations of student performance change \citep{cen2006learning} and recent attempts to replicate student behavior with LLMs \citep{kaser2024simulated}, AI2T and the Apprentice Learner master procedural academic tasks through rapid bottom-up induction from the limited instructional experiences available in automated tutoring systems. Beginning with limited prior knowledge, they learn domain expertise, often with 12 or fewer practice problems. Structure-mapping plays a critical role in enabling this data-efficient, human-like learning, which is many orders of magnitude faster than neural approaches like reinforcement learning and requires no external pretrained knowledge sources such as LLMs \citep{weitekamp2025decomposed}. Running these teachable simulated learners at scale \citep{weitekamp2025tutorgym}, and in real-time for applications where they are taught interactively from human instruction \citep{weitekamp2020CHI, weitekamp2024ai2t}, requires efficient and flexible subsystems for pattern matching and structure-mapping that work robustly over diverse environment representations, and which do not lag or break when applying the often complex or underconstrained knowledge structures induced by these simulated learners.

\textbf{The Cognitive Rule Engine (CRE)\footnote{https://github.com/DannyWeitekamp/Cognitive-Rule-Engine/},}
the toolset in which SMTB is implemented, was initially developed to meet the challenging computational needs of these simulated learners. Prior publications have reported on CRE's incremental pattern-matching algorithm CORGI \citep{weitekamp2025corgi}, which has quadratic first-match complexity guarantees that eliminate the exponential blowup of RETE-based and naive unification-based matching and beat well-optimized implementations such as SOAR \citep{laird2019soar} by orders of magnitude across all scales. Structure-mapping with Tight Bounds (SMTB) extends CRE with structure-mapping. We show that, like CORGI, SMTB is fast and flexible---it is 5--15x faster than SME on different problem sizes and works in domains with cycles instead of strictly tree-like nested predicate logic.  

CRE has grown into a broader project that aims to serve as a public open-source toolset for cognitive systems researchers. The aim is to accelerate cognitive system development, similar to how PyTorch and TensorFlow have accelerated deep learning research. CRE is implemented as a C++ framework that can be called from wrapper languages such as Python. Instead of specifying prior knowledge and domain content solely in a domain-specific language (e.g., PDDL or .soar files), CRE is designed with dynamic, first-class usage in mind: all data and knowledge structures, such as matching patterns and rules, can be flexibly specified directly in wrapper languages like Python. All objects created this way are fast C++ data structures; however, they are natively inspectable and editable within the user-facing wrapper language. This setup supports rapid development of learning mechanisms that treat knowledge as a flexibly inducible and modifiable material, rather than something that is only prespecified or recombined, as in speed-up learning approaches common in ACT-R and SOAR \citep{anderson1997act, laird2019soar}. Our hope is that this flexible framework-oriented design, combined with fast implementations of common algorithms like pattern matching and structure matching, will help facilitate rapid iteration in cognitive systems research and help scale beyond the scattered toy implementations and monolithic architecture-locked implementations (e.g., ACT-R, SOAR) that have dominated cognitive systems research for so long.



\section{The structure-mapping Engine (SME)}

\subsection{structure-mapping Theory} 
SME is built on Structure-Mapping Theory \citep{gentner1983structure}, which describes analogy formation as the process of establishing structural alignment between entities and expressions in a source and target domain. Structure-Mapping Theory imposes four main constraints on this process:

\begin{enumerate}
    \item \textbf{Structural Consistency}: The mapping must be one-to-one. E.g., if \texttt{Pressure}$\rightarrow$\texttt{Temp} then we cannot also have \texttt{Diameter}$\rightarrow$\texttt{Temp}.
    
    \item \textbf{Parallel Connectivity}: If two expressions correspond, then all of their arguments correspond. E.g., a commitment to \texttt{\mH{Flow}(\mA{beaker}, \mB{vial}, \mC{water}, \mD{pipe})}$\rightarrow$\texttt{\mH{Flow}(\mA{coffee}, \mB{icecube}, \mC{heat}, \mD{bar})} entails a commitment to \mA{beaker}$\rightarrow$\mA{coffee} and \mC{water}$\rightarrow$\mC{heat}, which would for instance forbid, \texttt{Liquid(\texttt{\mC{water}})}$\rightarrow\texttt{Liquid(\texttt{\mA{coffee}})}$.
    
    \item \textbf{Tiered Identicality}: Matches between same identity expressions e.g., \texttt{Flow}$\rightarrow$\texttt{Flow}, are prefferred over ones with different identities e.g., \texttt{Pressure}$\rightarrow$\texttt{Temp}. 

    \item \textbf{Systematicity}: (explicitly expressed here in two parts), (a) the preferred (higher scoring) mappings are those that align larger systems of relations, (b) especially ones consisting of higher-order relations with nested children. 
\end{enumerate}

\subsection{SME's Implementation} 
SME applies these four principles to its structure-mapping implementation \citep{forbus2017extending} in a three-phase process that relies heavily on a bias toward higher-order structure:

\textbf{Match hypothesis generation:} Enumerate all pairs of expressions in the source and target, storing pairs of match hypotheses that can plausibly match with one another. The rules for candidate plausibility are often configured on a domain-by-domain basis. Expressions must have the same arity (number of arguments), and constraints are set on which types of predicates can form correspondences. Each match hypothesis is assigned a weight reflecting its likelihood.  

\textbf{Kernel Formation:} Each root match hypothesis (ones that are not arguments of others) seeds the formation of kernels which are built up by collecting the mapping entailed by their descendants. Each kernel is essentially a partial mapping, and each is assigned a Structural Evaluation Score (SES) \citep{forbus1989structural} that aggregates the weights of its constituent match hypotheses in a manner that heavily favors kernels with deeper nested structures. 

\textbf{Kernel Merging: } Beginning with the highest-scoring kernel, kernels are greedily assembled \citep{forbus1990making} into complete matches by serially merging the highest-scoring kernel that is consistent with the match so far. Restarting the process from other seeds (lower-scoring initial kernels) yields a handful of alternative interpretations (three by default). 

As we discuss later, SMTB aligns with structure-mapping Theory and SME, along these four constraints, but is considerably more relaxed than SME on (4b). SMTB does not explicitly privilege mappings with deeper hierarchies, although it tends to find them anyway because deeper hierarchies tend to create larger systems of overlap. 

\section{When Privileging "Higher-Order" Correspondences can be a Failure Mode}

In our experience, we have not found that an explicit reliance on higher-order structure, such as SME kernels, is necessary for effective structure-mapping. Many routine situations require some form of associative partial alignment between representations, but rigidly privileging higher-order structures is either impossible, arbitrarily biased, or entirely incorrect. For instance:

\textbf{When there is no higher-order structure to exploit}, there is no way to bias structural mappings toward higher-order correspondences. This is not an uncommon situation. A broad survey of cognitive systems would likely find that flat first-order logic-like representations are more common than nested ones---not because flatness is preferable, but because hierarchy isn't always necessary. Classical planning, for instance, generally uses first-order representation languages (e.g., STRIPS/PDDL). Hierarchy is not a prerequisite for relational structure; as long as first-order expressions share entities, they have relational structure, and
structure-mapping ought to work in the absence of higher-order structure. However, in these cases SME kernels reduce to individual match hypotheses, removing any differentiating signal for kernel seeding and merging, and SME becomes equivalent to greedy merging over candidate expression pairs; a strategy unlikely to consistently find good mappings in larger problems.

\textbf{\textit{Higher-orderness} is often an arbitrary side effect of notation}. 
 In predicate logic notation, higher-order expressions encapsulate lower-order expressions. For instance, a second-order expression such as \texttt{Greater(Temp(coffee)}, 
 \texttt{Temp(icecube))} has first-order expressions \texttt{Temp(coffee)} and \texttt{Temp(icecube)} as arguments. However, we might reasonably question whether characterizing things as \textit{higher} or \textit{lower} solely on the basis of their referents ought to be given special treatment in determining structural connectedness. Predicate logic imparts \textit{higher-orderness} on objects as a side effect of its notation; this is, however, far from an exhaustive treatment of hierarchy. Predicate logic notation does not, for instance, treat predicates as \textit{higher} on the basis of derivational provenance, concept hierarchies, or part-whole relationships. \texttt{Father(fred, pops)} is equally first-order to \texttt{Parent(fred, pops)} and \texttt{IsMan(pops)}, despite the first being derivable from the others (from prior knowledge). Structure-mapping systems do not typically exploit these kinds of hierarchies, which are not captured by predicate logic's notation. And, perhaps they shouldn't. Our position is that relational connectedness---just the fact that objects are connected by some relation, whether nested or not---ought to be sufficient for establishing structural connectedness. SMTB, for instance, has no special reliance on higher-orderness, as SME does, allowing it to be fairly agnostic to data representation. SMTB will happily find good mappings over frame-based object-oriented ontologies, first-order predicates, and higher-order nested predicates all the same.  
 
\textbf{Relational cycles complicate which things are \textit{higher}.}
In a pure hierarchy (e.g., a tree of objects), directed relationships clearly mark objects as \textit{higher} or \textit{lower} than one another. Add a relation pointing in the opposite direction, and you have a cycle, and the clarity of which things are \textit{higher} is lost. Predicate logics often forbid this as a notational side effect, simply because they lack unique identifiers to specify higher things so they can be passed as arguments to lower ones. This is a representation limitation; it shouldn't be a limitation of structure-mapping.

\textbf{Higher correspondences should not require perfect child alignment}. SME kernels are consistent mappings between perfectly alignable expression trees. We have found this policy to be overly restrictive. In an ongoing project, we use SMTB-based structure-mapping to simulate students' ability to learn from step-by-step math solutions by self-explaining the underlying changes between lines \citep{shin2026sm_ast}. Our simulated learners use structure-mapping to find structural similarities between subsequent lines to abduce the transformation that occurred between them. Structure-mapping aligns the changed parts around the unchanged ones, simplifying the search for a consistent generalization that explains the observed transformation. For instance, consider the following pair of equation lines, parsed into abstract syntax trees expressed in predicate logic notation:  

\begin{table}[h]
\centering
\small
\begin{tabular}{l@{\hspace{4em}}l}
\toprule
\textbf{Base line} & \textbf{Target line} \\
\midrule
$4x+1 = 3^2$                                      & $4x+1 = 9$ \\
\midrule
\texttt{\mG{Equal}(}                              & \texttt{\mG{Equal}(} \\
\texttt{~~\mD{Add}(}                              & \texttt{~~\mD{Add}(} \\
\texttt{~~~~\mH{Mul}(\mA{Num(4)}, \mB{Sym(x)}),}  & \texttt{~~~~\mH{Mul}(\mA{Num(4)}, \mB{Sym(x)}),} \\
\texttt{~~~~\mC{Num(1)}),}                        & \texttt{~~~~\mC{Num(1)}),} \\
\texttt{~~\mF{Pow}(Num(3), Num(2)))}              & \texttt{~~\mF{Num(9)})} \\
\bottomrule
\end{tabular}
\caption{Color coded structure map of $4x+1 =3^2$ to $4x+1=9$ expressed in predicate logic.}
\label{tab:equations}
\label{tab:ast-example}
\end{table}

Both SME and SMTB succeed at mapping 4x+1 on the left-hand side of the two equations (indicated by \mD{\texttt{Add}}$\rightarrow$\mD{\texttt{Add}}). However, SME discards all other correspondences and thus fails to recognize that the only difference is localized to the right-hand side. An ideal mapping in this case should recognize that \mF{\texttt{Pow}}$\rightarrow$\mF{\texttt{Num(9)}} (this kind of task requires loose match hypotheses to permit possibilities like these). And that particular hypothesis should gain support by its shared placement on the right-hand side of the shared higher-order \mG{\texttt{Equal}}$\rightarrow$\mG{\texttt{Equal}}. However, since SME requires kernels to have valid argument correspondences all the way down, it cannot generate a kernel for \texttt{Equal} and thus misses the shared structure. The fact that the first-order expressions \texttt{Num(3)} and \texttt{Num(2)} have no counterparts in the target makes it impossible for SME to build a higher-order kernel. SMTB treats the higher-order structure more loosely, and thus avoids this failure mode.

\section{SMTB: structure-mapping with Tight Bounds}

The primary difference between SMTB and SME is that SMTB operates incrementally over a smaller unit of correspondence, assigning entities in the base to entities in the target at each search step, rather than assigning expressions to expressions or assigning sets of expressions collectively with kernels. A one-to-one entity assignment selects cells such that no assignment shares the same row or column. Table \ref{tab:sm-entities} shows the heat-flow analogy example, restated in terms of integer identifiers for each entity, an \textit{ identity mapping} (e.g. $0\rightarrow0, 1\rightarrow1,...$ etc.) over the first 7 entities in the base and target is the ideal mapping. Note that in this context we mean ``entities'' to include both expression predicates (e.g. \texttt{Greater$_1$}(4, 5)) and base entities (e.g. \texttt{beaker}).

Just like in SME, SMTB begins with a set of match hypotheses (e.g., 6: \texttt{Greater$_1$}(4, 5)$\rightarrow$ 6: \texttt{Greater}(4, 5) or 10: \texttt{Greater$_2$}(8, 9)$\rightarrow$ 6: \texttt{Greater}(4, 5)) between expressions. SMTB accepts different configuration options \texttt{args\_only}, \texttt{group}, or \texttt{pairwise} controlling how these hypotheses are interpreted as one or more \texttt{ExpressionPair} objects, with a weight $\phi_{k}$. 
For instance, in \texttt{group} mode (the default), a match hypothesis 10: \texttt{Greater$_2$}(8, 9)$\rightarrow$ 6: \texttt{Greater}(4, 5)) is interpreted as an \texttt{ExpressionPair} $[10,8,9]\rightarrow[6,4,5]$ which must be supported by $10\rightarrow6$, $8\rightarrow4$, and $9\rightarrow5$. In \texttt{arg\_only} the parent is excluded, and the resulting \texttt{ExpressionPair} is $[8,9]\rightarrow[4,5]$. In \texttt{pairwise} multiple \texttt{ExpressionPair} objects are created connecting the parent to each argument individually; in this case $[10,8]\rightarrow[6,4]$ and $[10,9]\rightarrow[6,5]$. Optionally, SMTB can be provided with independent candidate \texttt{EntityPairs} with weights $e_{i,j}$. SMTB's objective is to find a mapping $M(i)\rightarrow j$ that maximizes the $e_{i,j}$ and $\phi_k$ contributions of mutually consistent \texttt{ExpressionPairs}:

\begin{equation}
 S(M) = \sum_i e_{i,M(i)} + \sum_{T(k,i,M(i))} C(M,k)\phi_k   
 \label{eq:score}
\end{equation}
where $T(k,i,j)$ is an indicator function for when \texttt{ExpressionPair} $k$ has entity support for $i\rightarrow j$, and $C(M,k)$ is an indicator function accepting \texttt{ExpressionPairs} consistent with $M$. 

\begin{table}[t]
\centering
\small
\begin{tabular}{@{}r@{~~}l@{\hspace{1.2em}}r@{~~}l@{\hspace{2.2em}}r@{~~}l@{\hspace{1.2em}}r@{~~}l@{}}
\toprule
\multicolumn{4}{l@{\hspace{2.2em}}}{\textbf{Base: water flow}}
& \multicolumn{4}{l}{\textbf{Target: heat flow}} \\
\cmidrule(r{2.2em}){1-4} \cmidrule{5-8}
0: & \texttt{\mA{beaker}}            & ~8: & \texttt{Diameter(0)}
& 0: & \texttt{\mA{coffee}}          & 8: & \texttt{Liquid(0)} \\
1: & \texttt{\mB{vial}}              & ~9: & \texttt{Diameter(1)}
& 1: & \texttt{\mB{icecube}}         &    & \\
2: & \texttt{\mC{water}}             & 10: & \texttt{Greater$_2$(8, 9)}
& 2: & \texttt{\mC{heat}}            &    & \\
3: & \texttt{\mD{pipe}}              & 11: & \texttt{Cause(6, 7)}
& 3: & \texttt{\mD{bar}}             &    & \\
4: & \texttt{\mF{Pressure}(0)}       & 12: & \texttt{Liquid(2)}
& 4: & \texttt{\mF{Temp}(0)}         &    & \\
5: & \texttt{\mI{Pressure}(1)}       & 13: & \texttt{Flat-Top(2)}
& 5: & \texttt{\mI{Temp}(1)}         &    & \\
6: & \texttt{\mG{Greater$_1$}(4, 5)} &     &
& 6: & \texttt{\mG{Greater}(4, 5)}   &    & \\
7: & \texttt{\mH{Flow}(0, 1, 2, 3)}  &     &
& 7: & \texttt{\mH{Flow}(0, 1, 2, 3)} &   & \\
\bottomrule
\end{tabular}
\caption{The entities of Table~\ref{tab:sm-example}, numbered.}
\label{tab:sm-entities}
\vspace{-1em}
\end{table}

The main difficulty with incremental entity assignment is determining which assignments will successfully support the highest sum of \texttt{ExpressionPair} weights $\phi_k$ in an ideal match. An \texttt{ExpressionPair} is fully supported only when all of its supporting assignments have been made. Thus, it is difficult to know exactly which assignments will contribute to successfully locking in each $\phi_k$, and the scope of realizable \texttt{ExpressionPairs} is often unclear until several assignments later. Some heuristic $h(i,j)$ is needed to guide the choice of assignments $i \rightarrow j$ toward a good mapping. Ideally, this heuristic is \textit{admissible}; i.e., it is an upper bound on the score contribution of each assignment  \citep{pearl1984heuristics} in a completed mapping.

\subsection{Conceptual Framing with a Naive Upper Bound}

A naively admissible $h(i,j)$ is to assume that all candidate \texttt{ExpressionPairs} are realizable by distributing the weight of each $\phi_k$ across its $n_k$ supporting entity assignments. 

\begin{equation}
 h_{\textrm{naive}}(i,j) = e_{i,j} + \sum_{T(k,i,j)} C(M,k)\phi_k /n_k
\end{equation}

This is a very loose upper bound on the expected weight contribution of each assignment, because, in practice, many candidate \texttt{ExpressionPairs} are mutually exclusive. For instance, if we accept $[10,8,9]\rightarrow[6,4,5]$ then we cannot also accept $[6,4,5]\rightarrow[6,4,5]$, because the mapping must be one-to-one. Table \ref{tab:naive-matrix} shows this naive heuristic applied over a portion of the assignment matrix. The heatflow example is configured with 1.0 weight for each entity candidate pair, and $\phi_k=\frac{2}{3}n_k$ for $n_k>1$ and $\phi_k=0.5$ when $n_k=1$.\footnote{These particular weight choices aren't particularly important; they simply slightly favor non-unitary predicates when the weights are distributed, and larger predicates' weight isn't diluted upon distribution.} 

Table \ref{tab:naive-matrix} shows the naive heuristic applied at the first step of incremental backtracking search; thereafter, the heuristic matrix is recomputed after each assignment. With greedy backtracking search, the first assignment would choose $ 2\to 0$ with $h(2,0)=2.0$, a locally optimal but globally non-optimal choice. And after recomputing $h(i,j)$, it would zero out column 0 and row 2, since all $ 0\to j$ and $ i\to 2$ would be inconsistent. As the search continues, several complete non-optimal mappings are explored, eventually backtracking to the start, where $0\to 0$ would eventually lead to an optimal mapping. However, full backtracking search would continue searching after that until the entire space is exhaustively searched. 

\begin{table}[t]
\centering
\scriptsize
\setlength{\tabcolsep}{3pt}
\newcommand{\nv}[1]{\makebox[2.3em][c]{#1}}
\newcommand{\nc}{\nv{\color{black!30}--}}
\newcommand{\rowlab}[1]{\rotatebox[origin=c]{90}{\textbf{#1}}}
\begin{tabular}{@{}c@{\hspace{0.7em}}r@{\hspace{0.45em}}l@{\hspace{1.2em}}ccccccccc@{}}
\toprule
& & & \multicolumn{9}{c}{\textbf{Target: heat flow}} \\
\cmidrule(l){4-12}
& & & \nv{\mA{0}} & \nv{\mB{1}} & \nv{\mC{2}} & \nv{\mD{3}} & \nv{\mF{4}} & \nv{\mI{5}} & \nv{\mG{6}} & \nv{\mH{7}} & \nv{$\cdots$} \\
\midrule
\multirow{12}{*}{\rowlab{Base: water flow}}
& 0: & \texttt{\mA{beaker}}            & \nv{\textbf{1.67}} & \nv{1.00} & \nv{1.00} & \nv{1.00} & \nc & \nc & \nc & \nc & \nv{$\cdots$} \\
& 1: & \texttt{\mB{vial}}              & \nv{1.00} & \nv{\textbf{1.67}} & \nv{1.00} & \nv{1.00} & \nc & \nc & \nc & \nc & \nv{$\cdots$} \\
& 2: & \texttt{\mC{water}}             & \nv{\underline{2.00}} & \nv{1.00} & \nv{\textbf{1.67}} & \nv{1.00} & \nc & \nc & \nc & \nc & \nv{$\cdots$} \\
& 3: & \texttt{\mD{pipe}}              & \nv{1.00} & \nv{1.00} & \nv{1.00} & \nv{\textbf{1.67}} & \nc & \nc & \nc & \nc & \nv{$\cdots$} \\
& 4: & \texttt{\mF{Pressure}(0)}       & \nc & \nc & \nc & \nc & \nv{\textbf{1.67}} & \nv{1.00} & \nc & \nc & \nv{$\cdots$} \\
& 5: & \texttt{\mI{Pressure}(1)}       & \nc & \nc & \nc & \nc & \nv{1.00} & \nv{\textbf{1.67}} & \nc & \nc & \nv{$\cdots$} \\
& 6: & \texttt{\mG{Greater$_1$}(4, 5)} & \nc & \nc & \nc & \nc & \nc & \nc & \nv{\textbf{1.67}} & \nc & \nv{$\cdots$} \\
& 7: & \texttt{\mH{Flow}(0, 1, 2, 3)}  & \nc & \nc & \nc & \nc & \nc & \nc & \nc & \nv{\textbf{1.67}} & \nv{$\cdots$} \\
& 8: & \texttt{Diameter(0)}            & \nc & \nc & \nc & \nc & \nv{1.67} & \nv{1.00} & \nc & \nc & \nv{$\cdots$} \\
& 9: & \texttt{Diameter(1)}            & \nc & \nc & \nc & \nc & \nv{1.00} & \nv{1.67} & \nc & \nc & \nv{$\cdots$} \\
& 10: & \texttt{Greater$_2$(8, 9)}    & \nc & \nc & \nc & \nc & \nc & \nc & \nv{1.00} & \nc & \nv{$\cdots$} \\
& \multicolumn{2}{l}{~~$\vdots$}       & \nv{$\vdots$} & \nv{$\vdots$} & \nv{$\vdots$} & \nv{$\vdots$} & \nv{$\vdots$} & \nv{$\vdots$} & \nv{$\vdots$} & \nv{$\vdots$} & \nv{$\ddots$} \\
\bottomrule
\end{tabular}
\caption{$h_{naive}(i,j)$ for heat flow example. Ideal mapping bolded along diagonal. The \mC{\texttt{water}}$\rightarrow$ \mA{\texttt{coffee}} distractor (score \underline{2.00}) is best in its column, from \texttt{Liquid(\mC{water})}$\rightarrow$\texttt{Liquid(\mA{coffee})}.}
\label{tab:naive-matrix}
\vspace{-1em}
\end{table}


The heuristic value of any cell can change after any assignment, since some \texttt{ExpressionPairs} can be invalidated, which affects other rows and columns. This fact prevents us from viewing structure-mapping as equivalent to a Linear Assignment Problem (LAP)---a class of optimization problems with fast polynomial-time solutions, such as the Hungarian Algorithm \citep{kuhn1955hungarian}, where an optimal one-to-one mapping is found over a cost matrix of paired candidate mappings. LAP solvers will find consistent but naively greedy structure maps. Structure-mapping is closer to a Quadratic Assignment Problem (QAP) \citep{koopmans1957assignment}, which is known to be NP-Hard, and is often applied to optimal graph alignment problems \citep{conte2004thirty}. However, structure-mapping is arguably even harder than QAP since it uses arbitrary n-ary relations (instead of only binary ones that can be treated as graph edges), and higher-order relations---i.e., expressions can also be entities, so there is not always a clean node vs. edge distinction. 


This naive approach using $h_{naive}$ is wasteful on several accounts: 1) after each assignment it recomputes the heuristic by repeated summation instead of incremental updates, 2) its values reflect local relational connectivity instead of connectivity as part of larger connected systems, 3) it uses a bound which is far too loose because it sums the weights of \texttt{ExpressionPairs} which can never be realized simulatenously in the same one-to-one mapping, and 4) we have motivated it within an exhaustive backtracking search which would have $O(n!)$ runtime relative to problem size. The main features of SMTB fix issues 1 and 3 by tightening the heuristic and updating it incrementally. Issue 2 can be addressed separately by adding an extra term to the heuristic (described in Section 4.3) that weights total connectivity similarly to SME's Structure Evaluation Score (SES). With these features applied, issue 4 no longer becomes a problem because the heuristic is strong enough to find good matches greedily without backtracking.   

\subsection{Tightening the Bounds by Peeking Ahead}

SMTB provides a much tighter upper bound by explicitly accounting for the fact that each entity assignment $i\rightarrow j$ effectively forces some future assignment $u \rightarrow v$ to complete a subset of the \texttt{ExpressionPairs} that it began to fill in. We can formulate a tighter $h(i,j)_{tight}$ bound by pre-computing the maximal weight contribution of realizable $\phi_k$ supported by each $i\rightarrow j$. If we were to determine this value precisely, we would need to find the optimal mapping after pinning each $i\rightarrow j$, which is, of course, as hard as the original problem. Instead, we can make a first-order approximation that looks one step ahead of the current partial mapping $M$. Consider the set of \texttt{ExpressionPairs} in Table \ref{tab:peek-example} as an example.

\begin{table}[h]
\centering
\small
\setlength{\tabcolsep}{4pt}
\begin{tabular}{@{}l@{\hspace{0.8em}}l@{\hspace{2.5em}}l@{\hspace{0.8em}}l@{}}

\texttt{[0,1,2]}$\rightarrow$\texttt{[5,6,8]} & $\phi_0/n_k=1/3$ & \texttt{[0,1,4]}$\rightarrow$\texttt{[10,7,9]} & $\phi_4/n_k=2/3$ \\
\texttt{[0,1,2]}$\rightarrow$\texttt{[5,7,9]} & $\phi_1/n_k=1/3$ & \texttt{[0,2,4]}$\rightarrow$\texttt{[10,8,9]} & $\phi_5/n_k=1/3$ \\
\texttt{[0,3,4]}$\rightarrow$\texttt{[5,6,8]} & $\phi_2/n_k=1/3$ & & \\
\texttt{[0,3,4]}$\rightarrow$\texttt{[5,7,9]} & $\phi_3/n_k=1/3$ & & \\
\cmidrule(r){1-2}\cmidrule(l){3-4}
\multicolumn{2}{c}{\textbf{Candidate: $0\rightarrow 5$}} & \multicolumn{2}{c}
{\textbf{Candidate: $0\rightarrow 10$}} \\
\multicolumn{2}{c}{$h_{\textrm{naive}}(0,5)=4/3$} & \multicolumn{2}{c}{$h_{\textrm{naive}}(0,10)=3/3$} \\

\bottomrule
\end{tabular}
\caption{Example \texttt{ExpressionPairs} for two candidate assignments $0 \to 5$ and $0 \to 10$ with their corresponding naive heuristic values 4/3 and 3/3.}
\label{tab:peek-example}
\end{table}

The \texttt{ExpressionPairs} on the left and right are supported by the candidate assignments $0\rightarrow 5$ and $0\rightarrow 10$. Ignoring any $e_{i,j}$ contributions, the naive heuristic would assign $h(0,5)=4/3$ and $h(0,10)=3/3$, meaning it would choose $0 \rightarrow 5$ as the more favorable next match. However, no one-to-one mapping can simultaneously assign \texttt{[0,1,2]}$\rightarrow$\texttt{[5,6,8]} and \texttt{[0,1,2]}$\rightarrow$\texttt{[5,7,9]}, or \texttt{[0,3,4]}$\rightarrow$\texttt{[5,6,8]} and \texttt{[0,3,4]}$\rightarrow$\texttt{[5,7,9]}, so in reality a complete mapping commited to $0 \to 5$ would only support half of the weight $2/3$ suggested by the naive heuristic. It would have been better to select $0 \to 10$, which accumulates a weight of $3/3$ without contradiction. SMTB calculates a tighter heuristic by maintaining a dependency matrix for each $i \to j$ to estimate the maximum \texttt{ExpressionPair} weight that can be locked in while keeping a candidate assignment's dependent candidate assignments $u \to v$ consistently one-to-one.

\begin{table}[h]
\centering
\small
\setlength{\tabcolsep}{4pt}
\newcommand{\zz}{\textcolor{black!30}{--}}
\newcommand{\hdr}[1]{\multicolumn{1}{c}{\textbf{#1}}}
\begin{minipage}[t]{0.53\linewidth}
\centering
\textbf{Candidate: $0\rightarrow 5$}\\[3pt]
\begin{tabular}{@{}r@{\hspace{0.9em}}cccc@{\hspace{0.9em}}r@{}}
\toprule
& \hdr{6} & \hdr{8} & \hdr{7} & \hdr{9} & \hdr{$\mathbf{r_u}$} \\
\midrule
\textbf{1} & $1/6$ & \zz   & $1/6$ & \zz   & $1/6$ \\
\textbf{2} & \zz   & $1/6$ & \zz   & $1/6$ & $1/6$ \\
\textbf{3} & $1/6$ & \zz   & $1/6$ & \zz   & $1/6$ \\
\textbf{4} & \zz   & $1/6$ & \zz   & $1/6$ & $1/6$ \\
\midrule
\hdr{$\mathbf{c_v}$} & $1/6$ & $1/6$ & $1/6$ & $1/6$ & $\mathbf{2/3}$ \\
\bottomrule
\end{tabular}
\end{minipage}%
\begin{minipage}[t]{0.43\linewidth}
\centering
\textbf{Candidate: $0\rightarrow 10$}\\[3pt]
\begin{tabular}{@{}r@{\hspace{0.9em}}ccc@{\hspace{0.9em}}r@{}}
\toprule
& \hdr{7} & \hdr{9} & \hdr{8} & \hdr{$\mathbf{r_u}$} \\
\midrule
\textbf{1} & $1/3$ & \zz   & \zz   & $1/3$ \\
\textbf{4} & \zz   & $1/2$ & \zz   & $1/2$ \\
\textbf{2} & \zz   & \zz   & $1/6$ & $1/6$ \\
\midrule
\hdr{$\mathbf{c_v}$} & $1/3$ & $1/2$ & $1/6$ & $\mathbf{3/3}$ \\
\bottomrule
\end{tabular}
\end{minipage}
\caption{Dependency matrices $D$ for two candidate assignments.}
\label{tab:dep-matrices}
\end{table}

Each cell in a dependency matrix $D$ distributes \texttt{ExpressionPair} weight once more by another factor of $1/(n_k-1)$, since there are $(n_k-1)$ dependency slots for each \texttt{ExpressionPair}, and computes the row maximums $r_u = \max_{v} D_{u,v}$ and column maximums $c_v = \max_{u} D_{u,v}$. The final heuristic value is the minimum of the sums of those maximums:

\begin{equation}
 h(i,j)_{tight}=\min(\sum_u{r_u} ,{\sum_v{c_v})}    
\end{equation}

Table \ref{tab:dep-matrices} shows the dependency matrices for $0 \to 5$ and $0 \to 10$ and the tighter heuristic values $2/3$ and $3/3$, which in this case are exactly the maximum distributed weights achievable by each candidate assignment. The tight SMTB heuristic correctly prefers $0 \to 10$ over $0 \to 5$. 


\subsection{Preferring Global Structural Connectivity with Paired Walks}

The tight SMTB heuristic helps privilege assignments that maximize realizable structural connectivity locally, but carries no signal about the size of the total connected systems surrounding them. SME's Structural Evaluation Score (SES) evaluates connected system size through kernels realized via higher-order predicates. In first-order predicate encodings, like family trees encoded only with \texttt{Parent(child\_id, parent\_id)}, SES would provide no helpful signal. It couldn't help determine, for instance, if some family tree A was more structurally similar to trees B or C.
 
SMTB takes a simpler approach which uses paired walks to measure the shared connectivity around each $i \to j$ assignment. Setting \texttt{local\_walk\_depth} controls how many recursive steps $L$ are used to calculate a bias term $b_{i,j}$ reflecting the shared connectivity in neighboring entities. The base case $L=0$ adds $e_{i,j}$ to $b_{i,j}$, which includes some weight associated with entities sharing the same type (by default 1.0; less if they only share a parent type) plus a small amount of weight for any constant attribute values they have in common (.3 by default). At L=1 and greater, the walk progresses outward. Entity-valued attributes are checked for candidate entity map compatibility (i.e., a $e_{i,j}$ weight must be defined between them). Then their base weight $e_{i,j}$ is added to $b_{i,j}$, and if the depth limit has not been reached, they are added to the frontier so the walk can recurse and collect the pair weights of their neighbors, and so on. Walks grow geometrically with respect to their walk depth $L$; however, a covered set is maintained so no entity in the base or target is added to the frontier twice, bounding walk length to at most the domain size. 

Unlike SES, these paired walks can reflect shared connectivity even among first-order predicates. By default, they follow parent-argument relationships bidirectionally, so if entity/expression \texttt{Parent(fred, pops)} in the base domain takes entity \texttt{fred} as an argument, then a walk centered around \texttt{fred} can traverse backward through that connection and add \texttt{Parent(fred, pops)} to its frontier, assuming, of course, that the walk has paired them with entities in the target sharing an analogous parent-argument relationship. Unlike SES, this approach does not depend upon higher-order predicates to connect lower-order ones.

To align $b_{i,j}$ values with the smaller scale of $h(i,j)$, the bias matrix is normalized by it's maximum $\hat{b}_{i,j} = b_{i,j}/\max_{i,j}(b_{i,j})$, and given a fractional contribution $\alpha_d$ in the final heursitic, which decays exponentionally as a function of the current search depth $d$ (i.e. how many assignments have already been made). 

\setlength{\abovedisplayskip}{-1em}
\setlength{\belowdisplayskip}{1em}
\begin{align}
\alpha_d &= \alpha_0 e^{-d/\tau} \\
h(i,j)_{L=n} &= \frac{\alpha_d \hat{b}_{i,j} +h(i,j)_{tight}}{\alpha_d+1}   
\end{align}

The idea behind decaying the $\hat{b}_{i,j}$ contribution is that each $b_{i,j}$'s measure of system-level connectivity is more useful early on for selecting the initial assignments that seed larger connected groups. However, $b_{i,j}$ values are only calculated at start-up, so they become less helpful once some assignments have rendered portions of the candidate connections they accounted for unrealizable. $b_{i,j}$ also tend to be similar among connected entities, so they are more helpful for seeding group alignment than for maximally finishing aligning the constituent entities in a pair of connected systems. 

\section{Methods}
We evaluate SMTB on the SME corpus, a set of domains used in prior SME research \citep{forbus2017extending}, relative to SME v4 implemented in Allegro Common Lisp. We run both models head-to-head on the same device in a single thread, and compare the quality of the mappings they produce. The SME corpus consists of 5,845 base-target domain pairs with 5 sub-corpora \texttt{ap-physics} (1,137) \citep{klenk2009analogical}, \texttt{geometry} (873), \texttt{moraldm} (420) \citep{dehghani2008moraldm}, \texttt{oddity} (3,411) \citep{lovett2017modeling}, and \texttt{thermo} (4) \citep{forbus2001exploring}. Since SMTB and SME optimize over different objectives, we compare how each resulting mapping scores on both scales. The \textbf{SMTB objective} is $S(M)$ (equation \ref{eq:score}), the sum of all $e_{i,j}$ and $\phi_k$ contributions satisfied by each mapping. The \textbf{SME objective} is the sum of Structural Evaluation Scores (SES) of the realized match hypotheses. We normalize these scores by the value SME v4 achieves on each scale. 

Some mappings that SMTB can produce would be forbidden by SME. These include situations like the algebraic equations in Table \ref{tab:equations}, where higher-order predicates are mapped between the base and target that would have failed SME's kernel construction phase due to the absence of mappings in lower-order constituents. These \textit{kernel violations} contradict an assumption of SME's approach, but not the main assumptions of structure-mapping theory like \textit{Structural Consistency} (i.e., one-to-one mappings) and \textit{Parallel Connectivity}. We record the percentage of entity correspondences in SMTB's mappings that SME would consider \textbf{kernel violations}, and also report score results with those entity assignments removed from the final mapping.  

We configure SMTB to interpret corpus problems precisely as SME does. We translated SME corpus problems into JSON (so CRE's Python extension can read them) and set up SMTB to correctly interpret a wide variety of configuration options hard-coded into each problem description. Admittedly, this was tedious work, and language models helped with much of it. However, we verified that the starting match hypotheses generated by SMTB and SME were identical in this setup. Our timing measurements discount these translation steps (which are mostly written in Python). However, we measure the \textit{build} time associated with forming match hypotheses from the base and target domains after they have been initialized in SMTB's native format, as well as the \textit{search} time for each mapping. The SME v4 runtimes are analogous and rely on its built-in internal timers. 

SMTB is run with a bounded backtracking search parameterized by a branch width $w$ (how many candidate columns a row may try) and a branch depth $d$ (how deep the search may backtrack); $w{=}1, d{=}1$ is therefore purely greedy assignment with no backtracking at all. We also report a \texttt{LAP} baseline that solves a single linear assignment problem over the naive heuristic $h_{naive}$. 

\begin{figure}[tb]
\centering
\begin{subfigure}{.875\linewidth}
\centering
\includegraphics[width=\linewidth]{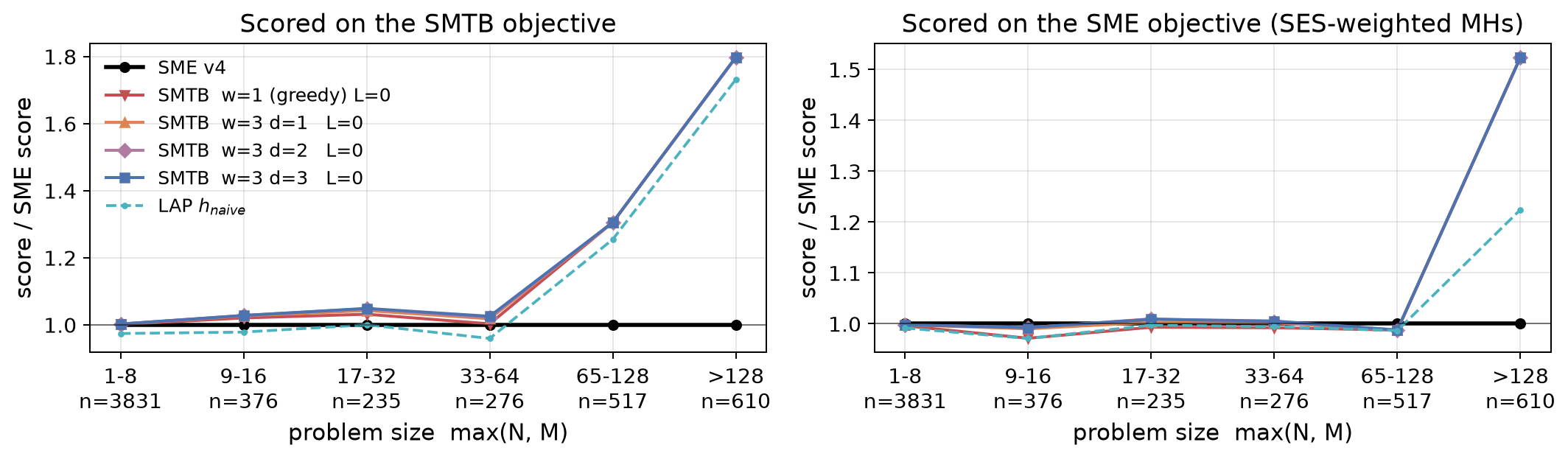}
\label{fig:score-vs-size}
\vspace{-1.2em}
\end{subfigure}
\begin{subfigure}{\linewidth}
\hspace*{.62cm} 
\includegraphics[width=.89\linewidth]{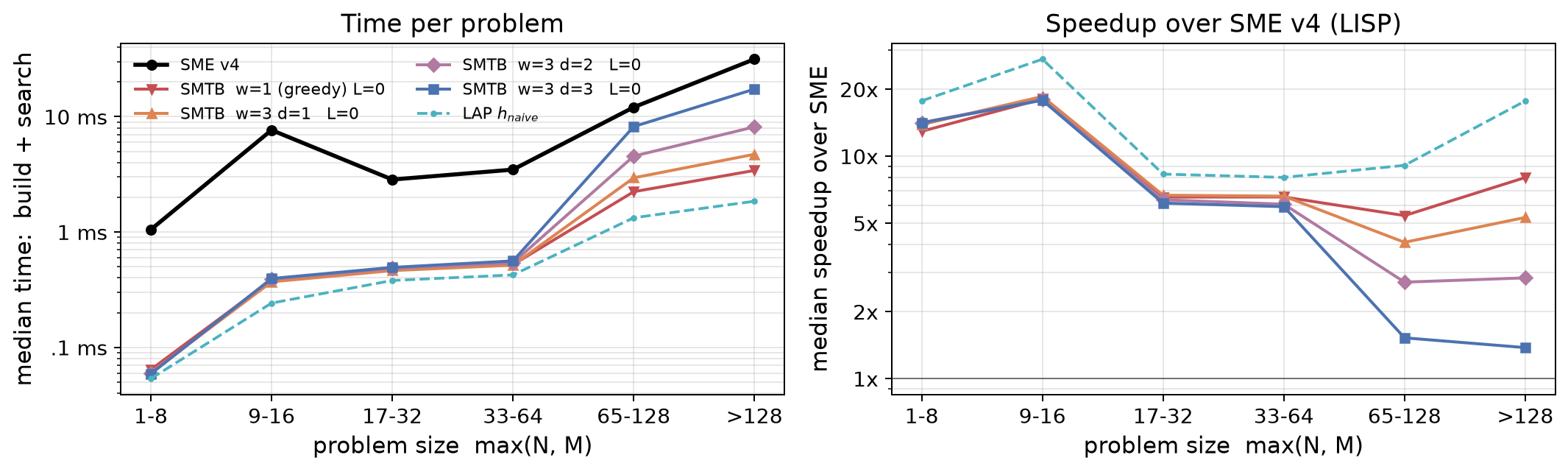}
\label{fig:time-vs-size}
\end{subfigure}
\vspace{-1.5em}
\caption{ Mapping quality ratio (top) under SMTB and SME's objectives relative to SME, and model solve time (bottom), including match hypothesis construction and search time. All results are graphed as a function of problem size (max number of entities between base and target).}
\label{fig:quality-and-cost-by-size}
\end{figure}

\section{Results}

The top two graphs of Figure \ref{fig:quality-and-cost-by-size} show SMTB's mapping score ratio relative to SME by problem size. For small problems (<64 entities), SMTB and SME perform nearly the same, with variants of SMTB showing at most $1.05\times$ better scores than SME evaluated on SMTB's objective and $0.99\times$ and $1.01\times$ SME's score evaluated on the SME objective. For medium problems with 65--128 entities, SMTB reaches $1.31\times$ on the SMTB objective, and for large problems with more than 128 entities it reaches $1.80\times$. For these large problems, SMTB also beats SME by $1.52\times$ on the SME objective. The LAP solver over $h_{naive}$ is consistently slightly worse than SMTB, but by all means a viable solution that also beats SME. 

The bottom two graphs of Figure \ref{fig:quality-and-cost-by-size} report total solution time over the whole SME corpus. SME takes 70.95\,s; purely greedy SMTB takes 5.97\,s and backtracking search with $w{=}3, d{=}3$ (for 27 total search arms) takes 30.99\,s. $w{=}3, d{=}1$ is a good tradeoff with a consistent 5-15x speedup over SME across problems sizes. This is also the configuration most similar to SME in terms of search branching, since, like SME, it chooses 3 different initial seed assignments.

\begin{table}[t]
\centering
\small
\setlength{\tabcolsep}{3pt}
\begin{tabular}{@{}lr@{\hspace{0.9em}}cccc@{\hspace{1.1em}}cccc@{\hspace{1.1em}}cccc@{}}
\toprule
& & \multicolumn{4}{c}{\textbf{\% Score gain vs.\ SME}} & \multicolumn{4}{c}{\textbf{\% Score gain vs.\ SME}} & \multicolumn{4}{c}{\textbf{\% Kernel}} \\
& & \multicolumn{4}{c}{(SMTB objective)} & \multicolumn{4}{c}{(SME objective $+$} & \multicolumn{4}{c}{\textbf{Violation}} \\
&&&&&& \multicolumn{4}{c}{Violations Removed)} &&&& \\
\cmidrule(lr){3-6}\cmidrule(lr){7-10}\cmidrule(l){11-14}
Domain & $n$ & $L{=}0$ & $L{=}1$ & $L{=}3$ & $L{=}5$ & $L{=}0$ & $L{=}1$ & $L{=}3$ & $L{=}5$ & $L{=}0$ & $L{=}1$ & $L{=}3$ & $L{=}5$ \\
\midrule
\texttt{ap-physics} & 1137 & 46.4 & 48.5 & 48.5 & 48.0 & 34.0 & 40.2 & 41.9 & 41.8 & 0.50 & 0.66 & 0.61 & 0.61 \\
\texttt{geometry}   &  873 & 0.8 & 1.0 & 0.8 & 0.9 & $-2.8$ & $-1.1$ & $-0.9$ & $-1.8$ & 5.41 & 6.27 & 5.33 & 5.67 \\
\texttt{moraldm}    &  420 & 3.0 & 3.5 & 3.4 & 3.3 & 1.2 & 1.5 & 1.4 & 1.6 & 0.14 & 0.14 & 0.14 & 0.00 \\
\texttt{oddity}     & 3411 & 0.5 & 0.3 & 0.5 & 0.6 & $-1.6$ & $-1.3$ & $-1.1$ & $-1.3$ & 3.61 & 5.84 & 4.17 & 4.39 \\
\texttt{thermo}     &    4 & 5.3 & 6.6 & 5.9 & 4.4 & $-6.0$ & $-1.3$ & 0.0 & 0.0 & 9.16 & 10.58 & 4.74 & 5.84 \\
\midrule
\textbf{ALL}        & 5845 & \textbf{21.6} & \textbf{22.5} & \textbf{22.6} & \textbf{22.4} & \textbf{10.3} & \textbf{14.7} & \textbf{16.0} & \textbf{15.9} & \textbf{1.24} & \textbf{1.78} & \textbf{1.38} & \textbf{1.44} \\
\bottomrule
\end{tabular}
\caption{Percent gain in mapping score over SME for SMTB $w{=}3, d{=}1$, varied by local walk depth $L=\{0,1,3,5\}$. Gains over SME are reported under SMTB's objective and under SME's objective. This time (unlike Figure \ref{fig:quality-and-cost-by-size}), for the SME objective, entity correspondences that SME kernels would have forbidden are removed. We report the percentage of these \textit{kernel violation} correspondences, which are small overall $1.24$-$1.78\%$, albeit as much as $10.58\%$ for some domains and configurations. }  
\label{tab:consistency-vs-L}
\end{table}

Adding paired walks (i.e., $L \geq 1$, see Table \ref{tab:consistency-vs-L}) marginally improves score quality, with a small cost to overall solve time. SMTB with $w{=}3, d{=}1$ scores $21.6\%$ above SME at $L=0$ and $22.6\%$ above SME at $L=3$, with a total corpus runtime: 8.47s for $L=0$, 8.63s for $L=3$, and 9.11s for $L=10$. Paired walks, in some cases, help SMTB yield mappings better aligned to SME's approach. For instance, Table \ref{tab:consistency-vs-L} also reports \textit{kernel violation} rates (right columns). When reporting SME's objective, we remove kernel violation mappings before scoring (middle columns). $L=5$ cuts those kernel violations from 9.16\% to 5.84\% in \texttt{thermo}, and from $0.14\%$ to $0.0\%$ in \texttt{moraldm}, and appreciably improves SME scores in \texttt{ap-physics}, from $34.0\%$ above SME at $L=0$ to $41.9\%$ above SME at $L=3$, albeit with a slight increase in kernel violations. 

\begin{table}[t]
\centering
\small
\setlength{\tabcolsep}{4pt}
\begin{tabular}{@{}lc@{\hspace{0.9em}}cc@{\hspace{1.1em}}cc@{\hspace{1.1em}}cc@{}}
\toprule
& \textbf{Depth} & \multicolumn{2}{c}{\textbf{Avg. \#}}
& \multicolumn{2}{c}{\textbf{Base Mapped}}
& \multicolumn{2}{c}{\textbf{Exprs. Mapped}} \\
\cmidrule(lr){3-4}\cmidrule(lr){5-6}\cmidrule(l){7-8}
Domain &   & Base & Expr. & SMTB & SME & SMTB & SME \\
\midrule
\texttt{ap-physics} & 7.1 & 47.7 & 100.2 & \textbf{20.1} & 13.5 & \textbf{23.0} & 13.3 \\
\texttt{geometry}   & 1.4 &  2.3 &   1.7 & 2.1 & 2.1 & 1.3 & 1.3 \\
\texttt{moraldm}    & 4.0 & 16.1 &  17.2 & \textbf{15.0} & 9.3 & 1.9 & 1.8 \\
\texttt{oddity}     & 1.5 &  2.4 &   1.8 & 2.4 & 2.4 & 1.7 & 1.7 \\
\texttt{thermo}     & 7.5 & 10.8 & 134.8 & 9.0 & 7.8 & 69.8 & 61.8 \\
\midrule
\textbf{ALL}        & \textbf{2.7} & \textbf{12.2} & \textbf{22.2} & \textbf{6.7} & \textbf{5.0} & \textbf{5.9} & \textbf{3.9} \\
\bottomrule
\end{tabular}

\caption{Structure of each corpus domain, and how many base entities (e.g. \texttt{coffee}) and expressions (e.g. \texttt{Liquid(coffee)}) are mapped by SMTB and SME. In domains with more nested higher-order structure like \texttt{ap-physics} and \texttt{moraldm}, SMTB with $w{=}3, d{=}1, L{=}0$ tends to find mappings that include about 50\% more mutually consistent correspondences.}
\label{tab:domain-structure}
\end{table}

Table \ref{tab:consistency-vs-L} also shows that most of SMTB's overall advantage under the SME objective comes from domains where SME hasn't already saturated the available entities. SMTB has the largest gains in the \texttt{ap-physics} domain, with smaller benefits in \texttt{moraldm} and small losses or equivalence elsewhere in the more saturated domains. Table \ref{tab:domain-structure} shows that the gains in \texttt{ap-physics} and \texttt{moraldm} are from SMTB finding mappings that include larger consistent paired systems of base entities and expressions. SMTB's objective is directly proportional to the number of correspondences, but SME's SES-based objective heavily favors systems with more higher-order structure (as opposed to just big ones). The fact that SMTB beats SME under the SME objective in these larger nested domains verifies that SMTB is not scoring higher as a result of pursuing a different objective, or from allowing more relaxed mappings (like our algebra example in Table \ref{tab:equations}). Kernel violation removal cuts out all relaxed correspondences like these. This leaves us with the conclusion that SMTB is indeed better than SME at maximizing the SME objective, and in general, at forming good mappings in large domains with deep hierarchical structure.

\section{Discussion}

We have promoted SMTB largely on practical grounds: it is implemented in a fast, Python-accessible toolkit, it works flexibly across different representational choices, and it finds good mappings in large problems (>128 entities) with higher-order structure. With respect to theory, SMTB diverges from structure-mapping Theory and SME on a few small details: 1) whether or not higher-orderness should be privileged explicitly (point 4b in Section 2.1); and 2) whether mapping correspondences are formed serially, entity-by-entity, or by groups of higher-order connected systems (i.e. kernels). For 1) SMTB offers evidence that good higher-order mappings can be found without giving them special treatment. Perhaps then, a theory of structural alignment based only on connectedness is sufficient; it certainly admits more flexible correspondences, which we've found necessary in some cases (e.g. the algebra example). For 2) SMTB assigns correspondences over smaller mental units. Future human experiments---if they could deconfound a higher-orderness preference from a preference for larger connected systems---may be able to clarify where humans fall along these points. 

\section{Conclusion}
We are, however, more excited by the practical trajectory of CRE and SMTB. Cognitive systems offer powerful untapped affordances for artificial intelligence, since they tend to solve precisely the problems that riddle naive data-driven neural network-based approaches like LLMs. Symbolic cognitive systems often learn considerably faster than neural approaches like reinforcement learning and typically induce interpretable, and robust knowledge structures. We should be reminded that the meteoric improvements in AI capabilities over just the last 5 years are largely the product of massively scaling a narrow AI paradigm. The research preceding that scaling was accelerated by good tools (e.g., PyTorch \& Tensorflow). It is not unreasonable to believe that cognitive systems could experience a similar trajectory, if the right tools set the stage for them to scale in the same way. Structure-mapping would play a key role in such a revolution since it is a key component of powering rapid human-like learning and flexible pattern matching capabilities with over symbolic knowledge representations.

\vspace{-0.1in}

{\parindent -10pt\leftskip 10pt\noindent
\bibliographystyle{cogsysapa}
\bibliography{references}

}

\appendix


\end{document}